\documentclass[runningheads]{llncs}
\usepackage[T1]{fontenc}
\usepackage{graphicx,verbatim}
\usepackage{amsmath}
\usepackage{algorithm}
\usepackage{algorithmic}
\usepackage{booktabs}
\usepackage{multirow}
\usepackage{cite}
\usepackage{xcolor}
\usepackage[table]{xcolor}  % Add this to preamble
\begin{document}
\title{VGS-Decoding: Visual Grounding Score Guided Decoding for Hallucination Mitigation in Medical VLMs}
%\titlerunning{Abbreviated paper title}
% If the paper title is too long for the running head, you can set
% an abbreviated paper title here
%
\begin{comment}  %% Removed for anonymized MICCAI submission
\author{First Author\inst{1}\orcidID{0000-1111-2222-3333} \and
Second Author\inst{2,3}\orcidID{1111-2222-3333-4444} \and
Third Author\inst{3}\orcidID{2222--3333-4444-5555}}
%
\authorrunning{F. Author et al.}
% First names are abbreviated in the running head.
% If there are more than two authors, 'et al.' is used.
%
\institute{Princeton University, Princeton NJ 08544, USA \and
Springer Heidelberg, Tiergartenstr. 17, 69121 Heidelberg, Germany
\email{lncs@springer.com}\\
\url{http://www.springer.com/gp/computer-science/lncs} \and
ABC Institute, Rupert-Karls-University Heidelberg, Heidelberg, Germany\\
\email{\{abc,lncs\}@uni-heidelberg.de}}

\end{comment}

\author{
 Govinda Kolli$^{1,2}$, Adinath Madhavrao Dukre$^{1}$, Behzad Bozorgtabar$^{3}$,  \\ Dwarikanath Mahapatra$^{4}$, Imran Razzak$^{1}$ }
 \institute{
$^{1}$MBZUAI, Abu Dhabi, UAE \\
$^{2}$Rajiv Gandhi University of Knowledge Technologies, India \\
$^{3}$ Aarhus University, Aarhus, Denmark\\
$^{4}$ Khalifa University, Abu Dhabi, UAE\\
}

\maketitle              % typeset the header of the contribution
\begin{abstract}
Medical Vision-Language Models (VLMs) often hallucinate by generating responses
based on language priors rather than visual evidence, posing risks in clinical
applications. We propose Visual Grounding Score Guided Decoding
(VGS-Decoding), a training-free method to mitigate hallucinations during
inference. Our key insight is that hallucinated tokens maintain or increase their
probability when visual information is degraded, while visually grounded tokens
decrease in probability. We introduce the Visual Grounding Score (VGS), which measures each token's visual dependency by comparing distributions from
original and distorted images. During decoding, we reweight probabilities by
amplifying visually grounded tokens while suppressing hallucinations. Unlike
fixed-weight contrastive methods, VGS-Decoding provides per-token adaptive
control. Experiments on MIMIC-Diff-VQA and VQA-RAD across LLaVA-Med,
CheXagent, and MedGemma demonstrate consistent improvements, with up to $+9.12\%$ overall gain and $+8.98\%$ in open-ended 
recall, while introducing only $2\times$ inference overhead and 
no additional training, making it practical for clinical deployment. Upon acceptance, code will be released publicly to facilitate 
reproducibility.

\keywords{Medical VQA \and Hallucination Mitigation \and Vision-Language Models \and Visual Grounding \and Contrastive Decoding}
% Authors must provide keywords and are not allowed to remove this Keyword section.

\end{abstract}
\section{Introduction}
\label{sec:intro}

Medical Vision-Language Models (VLMs) have emerged as powerful tools for clinical applications, demonstrating remarkable capabilities in tasks such as visual question answering (VQA), radiology report generation, and diagnostic assistance~\cite{li2023llava,chen2024chexagent,yang2024medgemma}. Models like LLaVA-Med~\cite{li2023llava}, trained on large-scale biomedical image-text pairs, have shown promising performance on medical VQA benchmarks including VQA-RAD~\cite{lau2018dataset} and MIMIC-Diff-VQA\cite{hu2023expert}. Despite these advances, medical VLMs remain vulnerable to hallucinations, generating clinically plausible but factually incorrect responses that are unsupported by the visual evidence~\cite{wu2024hallucination,gu2025medvh}. In high-stakes medical scenarios, such hallucinations pose significant risks, potentially leading to misdiagnosis and inappropriate clinical decisions, and dangerous automation bias~\cite{zheng2025large}.

% Hallucinations in VLMs primarily arise from the model's over-reliance on language priors rather than visual information~\cite{goyal2017making,leng2024vcd, huang2024opera}. When generating responses, models may draw upon statistical patterns learned during training, producing plausible-sounding answers that fail to reflect the actual content of the medical image. This is particularly problematic in medical VQA, where precise identification of anatomical locations, pathological findings, and diagnostic details is critical.

Hallucinations in VLMs primarily arise from over-reliance on 
language priors rather than visual information~\cite{goyal2017making,
leng2024vcd,huang2024opera}, where models draw upon statistical 
patterns learned during training. This is 
particularly problematic in medical VQA, where precise 
identification of anatomical locations, pathological findings, 
and diagnostic details is critical.
Several decoding-time strategies have been proposed to mitigate hallucinations in general-domain VLMs, building upon foundational contrastive decoding techniques in NLP \cite{li2022contrastive}. Visual Contrastive Decoding (VCD)~\cite{leng2024vcd} contrasts output distributions from original and noise-distorted images using a subtractive formula in log-space. Decoding by Contrasting
Layers(DoLA)~\cite{chuang2024dola} exploits the hierarchical encoding of factual knowledge in transformer layers by contrasting early and late layer outputs. Over-trust
Penalty and a Retrospection-Allocation(OPERA)~\cite{huang2024opera} introduces a penalty term during beam search to address over-trust in summary tokens. While these methods have shown effectiveness on general VQA benchmarks like POPE~\cite{li2023pope} and MME~\cite{fu2023mme}, we find that they often fail on medical VQA, sometimes degrading performance by over 6\% compared to baseline greedy decoding. This failure stems from the unique characteristics of medical VQA: shorter responses, specialized vocabulary, and the critical importance of precise anatomical and pathological terms where aggressive correction can be harmful.

In this paper, we propose Visual Grounding Score Guided Decoding (VGS-Decoding) , a training-free method specifically designed for hallucination mitigation in medical VQA. Our key insight is that hallucinated tokens exhibit a distinctive signature: when visual information is degraded, hallucinated tokens maintain or \textit{increase} their probability (being driven by language priors), while visually grounded tokens \textit{decrease} in probability. We formalize this observation through the Visual Grounding Score (VGS), a novel metric that quantifies each token's visual dependency
% \begin{equation}
% \text{VGS}(t) = \frac{P_{\text{orig}}(t) - P_{\text{dist}}(t)}{P_{\text{orig}}(t) + P_{\text{dist}}(t)}
% \end{equation}
% where $P_{\text{orig}}(t)$ and $P_{\text{dist}}(t)$ are the token probabilities conditioned on original and distorted images, respectively.
Unlike VCD's fixed-weight subtractive approach, VGS provides a bounded, interpretable measure in $[-1, +1]$ that enables \textit{per-token adaptive} correction through multiplicative reweighting.
% \begin{equation}
% P_{\text{final}}(t) \propto P_{\text{orig}}(t) \times (1 + \alpha \cdot \text{VGS}(t))
% \end{equation}

This formulation offers several advantages: (1) tokens with high VGS (visually grounded) are amplified, (2) tokens with negative VGS (potential hallucinations) are suppressed, and (3) neutral tokens remain largely unchanged, preserving the model's linguistic coherence.

We evaluate VGS-Decoding on two medical VQA benchmarks (VQA-RAD~\cite{lau2018dataset} and MIMIC-Diff-VQA~\cite{hu2023expert} ) across three diverse medical VLMs (LLaVA-Med, CheXagent~\cite{chen2024chexagent}, and MedGemma~\cite{yang2024medgemma}). Our experiments reveal that VGS-Decoding is the only method that consistently improves all models, achieving up to +9.12\% overall improvement. In contrast, existing methods (VCD, DoLA, OPERA) frequently degrade performance on medical VQA, highlighting the need for domain-specific solutions.
 Our main contributions are:
\begin{itemize}
    \item[$\bullet$]  We identify that existing hallucination mitigation methods often fail on medical VQA, motivating the need for domain-adapted approaches.
    \item[$\bullet$]  We propose the Visual Grounding Score (VGS), a novel metric that quantifies token-level visual dependency through normalized probability difference.
    \item[$\bullet$]  We introduce VGS-Decoding, a training-free method that achieves consistent improvements across diverse medical VLMs and datasets.
\end{itemize}

\section{Method}
\label{sec:method}

\begin{figure}[!htb]
\centering
\includegraphics[width=\textwidth]{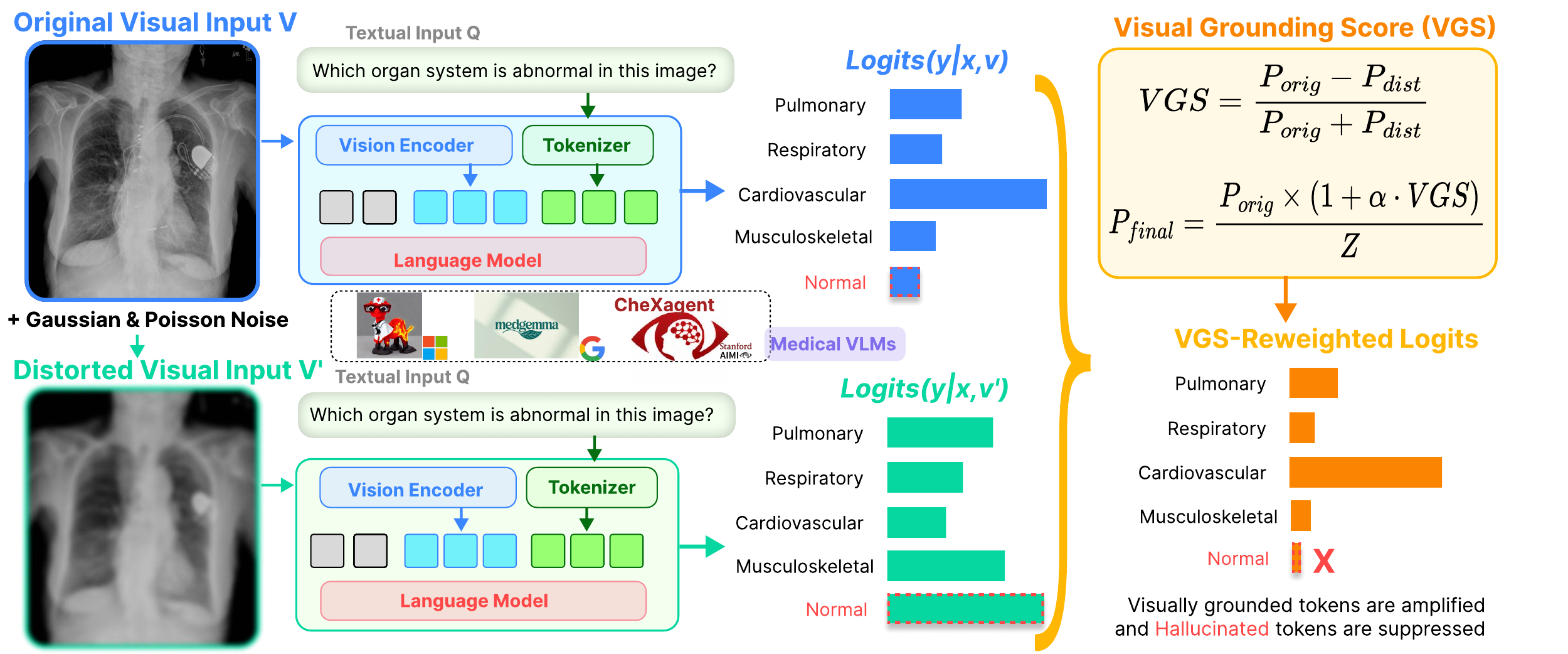}
\caption{Overview of VGS-Decoding. Both the original image $V$ and 
its noise-distorted version $V'$ are passed through the medical VLM, 
producing token distributions $P_{\text{orig}}$ (blue) and 
$P_{\text{dist}}$ (green). The Visual Grounding Score (VGS) is 
computed as their normalized probability difference and used for 
multiplicative reweighting: visually grounded tokens (positive VGS) 
are amplified, while hallucinated tokens (negative VGS, 
\textcolor{red}{$\times$}) are suppressed.}
\label{fig:Overview}
\end{figure}

We present Visual Grounding Score Guided Decoding (VGS-Decoding), a training-free approach for mitigating hallucinations in medical VLMs. Our method operates during inference by adaptively reweighting token probabilities based on their visual grounding score. Fig~\ref{fig:Overview} illustrates the overall framework.
%%%%%%%%%%%%%%%%%%%%%%%%%%%%%%%%%%%%
%%%%%%%%%%%%%%%%%
\subsection{Preliminaries}
\label{sec:preliminaries}
%%%%%%%%%%%%%%%%%%%%%%%%%%%%%%%%%%%%%%
% Given a medical image $V$ and a text query $Q$, a vision-language model generates a response token-by-token in an autoregressive manner. At each decoding step $t$, the model produces a probability distribution over the vocabulary:
% \begin{equation}
% P_{\text{orig}}(y_t \mid V, Q, y_{<t}) = \text{softmax}(\mathbf{z}_t)
% \end{equation}
% where $\mathbf{z}_t$ denotes the logits and $y_{<t}$ represents previously generated tokens. Standard greedy decoding selects $y_t = \arg\max P_{\text{orig}}(y_t)$.
Given a medical image $V$ and a text query $Q$, a vision-language model generates a response token-by-token in an autoregressive manner. At each decoding step $t$, the model produces a probability distribution over the vocabulary:
\begin{equation}
    P_{orig}(y_t | V, Q, y_{<t}) = \text{softmax}(z_t)
\end{equation}
where $z_t$ denotes the logits and $y_{<t}$ represents previously generated tokens. Standard greedy decoding selects $y_t = \arg\max_{y} P_{orig}(y)$.

Hallucinations occur when the model generates tokens based on language priors rather than visual evidence~\cite{rohrbach2018object,favero2024multi}. For instance, when asked about the location of a finding in a chest X-ray, a model might generate ``left'' based on statistical co-occurrence patterns even when the image shows a right-sided abnormality.

\subsection{Visual Grounding Score}
\label{sec:vgs}

Our method is motivated by a key observation: visually grounded tokens and hallucinated tokens exhibit distinct behaviors when visual information is degraded. Specifically, visually grounded tokens show decreased probability when the image is corrupted, as the model loses access to the supporting visual evidence. In contrast, hallucinated tokens maintain or even increase their probability under visual corruption, since these tokens are driven by language priors rather than actual visual content.

To exploit this behavioral signature, we first construct a distorted version of the input image $V' = \mathcal{D}(V)$ by applying a combination of Gaussian and Poisson noise~\cite{hendrycks2019benchmarking,LiaZeh_VisionAmplified_MICCAI2025,bushberg2011essential}:
\begin{equation}
V' = V + \mathcal{N}(0, \sigma^2) + \text{Poisson}(\lambda)
\end{equation}
where $\sigma$ and $\lambda$ control the noise intensity. We then obtain the probability distribution conditioned on the distorted image:
\begin{equation}
P_{\text{dist}}(y_t \mid V', Q, y_{<t}) = \text{softmax}(\mathbf{z}'_t)
\end{equation}

We define the Visual Grounding Score (VGS) as the normalized difference between the original and distorted distributions:
\begin{equation}
\text{VGS}(t) = \frac{P_{\text{orig}}(t) - P_{\text{dist}}(t)}{P_{\text{orig}}(t) + P_{\text{dist}}(t)}
\label{eq:vgs}
\end{equation}

The VGS metric offers several desirable properties. First, it is bounded within $[-1, +1]$, ensuring stable and interpretable values throughout the decoding process. A positive VGS indicates that the token is visually grounded, as its probability decreases when visual information is degraded. Conversely, a negative VGS suggests potential hallucination, where the token's probability increases or remains stable without proper visual support. A VGS near zero indicates visual independence, typical of function words that are equally likely regardless of image quality.

\subsection{VGS-Guided Decoding}
\label{sec:decoding}

Using the VGS metric, we reweight the original probability distribution to amplify visually grounded tokens while suppressing potential hallucinations:
\begin{equation}
P_{\text{final}}(t) \propto P_{\text{orig}}(t) \times \max(1 + \alpha \cdot \text{VGS}(t), \delta)
\label{eq:reweight}
\end{equation}
where $\alpha > 0$ controls the reweighting strength and $\delta > 0$ is a small constant (default $\delta = 0.01$) to prevent zero probabilities. The final distribution is normalized to sum to one.

The multiplicative factor $(1 + \alpha \cdot \text{VGS}(t))$ provides adaptive correction based on each token's visual grounding. When $\text{VGS}(t) > 0$, the factor exceeds one, amplifying visually grounded tokens. When $\text{VGS}(t) < 0$, the factor falls below one, suppressing potentially hallucinated tokens. When $\text{VGS}(t) \approx 0$, the factor remains close to one, preserving neutral tokens such as function words. This per-token adaptive mechanism ensures that correction strength is proportional to the degree of visual grounding or hallucination.

\subsection{Algorithm}
\label{sec:algorithm}

Algorithm~\ref{alg:vgs} summarizes the complete VGS-Decoding procedure. The image is distorted once before decoding begins. At each decoding step, we compute probability distributions from both the original and distorted images, calculate the VGS for each vocabulary token, apply the multiplicative reweighting, and select the token with maximum adjusted probability.

\begin{algorithm}[t]
\caption{VGS-Decoding}
\label{alg:vgs}
\begin{algorithmic}[1]
\REQUIRE Image $V$, query $Q$, model $\mathcal{M}$, strength $\alpha$, noise params $(\sigma, \lambda)$
\ENSURE Generated response $Y$
\STATE $V' \leftarrow V + \mathcal{N}(0, \sigma^2) + \text{Poisson}(\lambda)$
\STATE $Y \leftarrow [\;]$
\WHILE{not end-of-sequence}
    \STATE $P_{\text{orig}} \leftarrow \mathcal{M}(V, Q, Y)$
    \STATE $P_{\text{dist}} \leftarrow \mathcal{M}(V', Q, Y)$
    \FOR{each token $t$ in vocabulary}
        \STATE $\text{VGS}(t) \leftarrow \frac{P_{\text{orig}}(t) - P_{\text{dist}}(t)}{P_{\text{orig}}(t) + P_{\text{dist}}(t)}$
        \STATE $P_{\text{final}}(t) \leftarrow P_{\text{orig}}(t) \times \max(1 + \alpha \cdot \text{VGS}(t), \delta)$
    \ENDFOR
    \STATE $P_{\text{final}} \leftarrow P_{\text{final}} / \sum P_{\text{final}}$
    \STATE $y \leftarrow \arg\max P_{\text{final}}$
    \STATE $Y \leftarrow Y \cup \{y\}$
\ENDWHILE
\RETURN $Y$
\end{algorithmic}
\end{algorithm}

\subsection{Computational Complexity}
\label{sec:complexity}

VGS-Decoding requires two forward passes per decoding step, one for the original image and one for the distorted image, resulting in approximately twice the inference time of standard greedy decoding. This computational cost is comparable to VCD and significantly lower than OPERA, which requires beam search with multiple candidates. The image distortion is performed once before decoding begins and adds negligible overhead.

\section{Experiments}
\label{sec:experiments}

\subsection{Datasets}
\label{sec:datasets}

We evaluate VGS-Decoding on two medical VQA benchmarks. VQA-RAD~\cite{lau2018dataset} contains 315 radiology images including CT, MRI, and X-ray with 3,515 question-answer pairs. Following standard practice, we use the test set comprising 451 questions with 200 open-ended and 251 closed-ended samples, covering diverse clinical queries about anatomical structures, abnormalities, and imaging modalities. MIMIC-Diff-VQA~\cite{hu2023expert} is derived from the MIMIC-CXR database~\cite{johnson2019mimic} and focuses on longitudinal chest X-ray analysis with question-answer pairs about differences between sequential radiographs. We evaluate on 13,121 test samples with 8,380 open-ended and 4,741 closed-ended questions to assess generalization to larger-scale medical VQA.

\subsection{Models}
\label{sec:models}

We evaluate three diverse medical VLMs to demonstrate generalizability. LLaVA-Med~\cite{li2023llava} is a 7B parameter model adapted from LLaVA through curriculum learning on biomedical image-text pairs from PubMed Central, representing general-purpose medical VLMs. CheXagent~\cite{chen2024chexagent} is a 3B parameter model specifically trained on MIMIC-CXR's official training set~\cite{johnson2019mimic} for chest X-ray interpretation, representing domain-specialized medical VLMs with strong in-domain performance. MedGemma~\cite{yang2024medgemma} is a 4B parameter model from Google trained on diverse medical imaging data, representing efficient smaller-scale medical VLMs.

\subsection{Baselines}
\label{sec:baselines}

We compare VGS-Decoding against four decoding strategies. Greedy decoding selects the highest probability token at each step, serving as the baseline without any hallucination mitigation. VCD~\cite{leng2024vcd} contrasts distributions from original and noise-distorted images using a subtractive log-space formula with recommended settings of $\alpha=1.0$ and diffusion noise. DoLA~\cite{chuang2024dola} contrasts early and late transformer layers to surface factual knowledge using dynamic layer selection as recommended. OPERA~\cite{huang2024opera} applies over-trust penalty and retrospection-allocation during beam search decoding with beam size 5, though it is computationally expensive and less effective on short-length generations typical in medical VQA~\cite{liu2021slake}, as beam search is known to suffer from length bias and degradation in open-ended generation tasks \cite{meister2020if}.

\noindent\textbf{Evaluation Metrics.}
Following standard medical VQA evaluation protocols~\cite{Antol_2015_ICCV,li2023llava, lau2018dataset}, we report three metrics. Closed-ended accuracy measures the percentage of correctly answered yes/no and multiple-choice questions, evaluating precise clinical judgment. Open-ended recall computes token-level recall between generated and ground-truth answers, measuring coverage of key medical terms. Overall accuracy provides a weighted combination based on question type distribution.
\noindent\textbf{Implementation Details}
For VGS-Decoding, we apply Gaussian noise with $\sigma=0.07$ and Poisson noise with $\lambda=70$ for image distortion. The reweighting strength is set to $\alpha=1.0$ as default across all models for fair comparison. All experiments use greedy decoding. Inference is performed on NVIDIA A100 GPUs. Statistical significance is computed using McNemar's test~\cite{mcnemar1947note} with 10,000 bootstrap samples~\cite{dror-etal-2018-hitchhikers}.

\subsection{Results}
\label{sec:results}

\subsubsection{Main Results}

\begin{figure}[t]
\centering
\includegraphics[width=\textwidth]{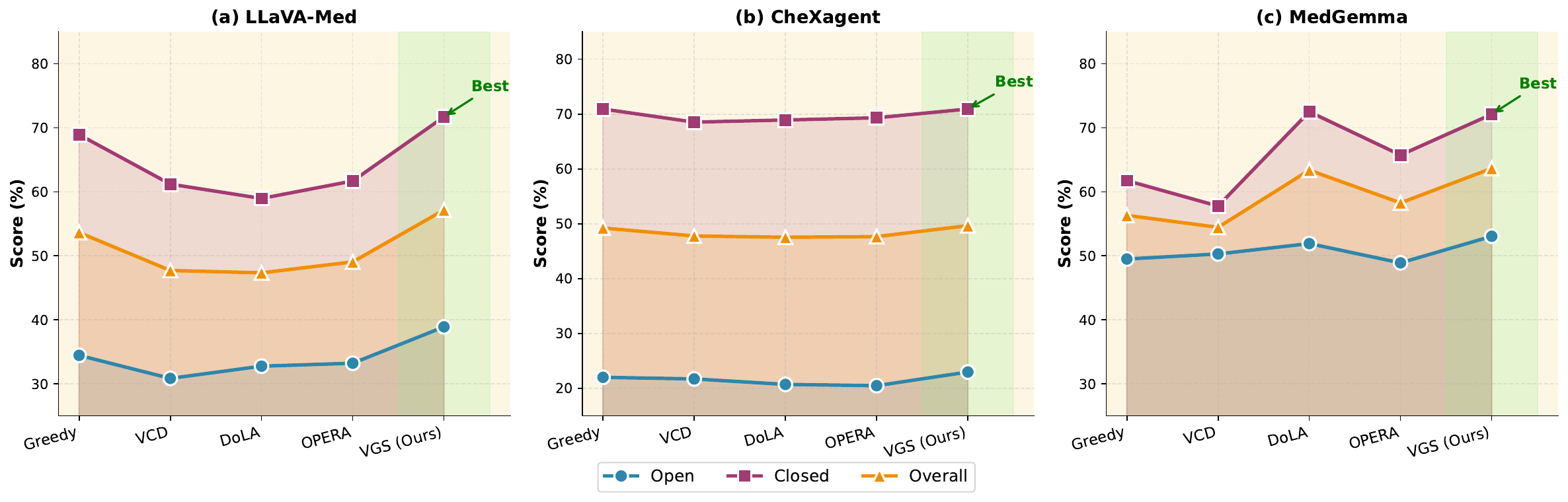}
\caption{Decoding method comparison on VQA-RAD. VGS-Decoding consistently outperforms baselines across all three medical VLMs.}
\label{fig:results}
\end{figure}

Table~\ref{tab:main_results} and Fig~\ref{fig:results} presents the comparison across all models and datasets. VGS-Decoding achieves consistent improvements across experimental settings, improving all three models on both datasets and achieving best results on 6 out of 6 model-dataset combinations. Existing methods frequently degrade performance on medical VQA, with VCD showing up to $-5.93\%$ and OPERA up to $-5.25\%$ drops. The largest improvements are observed on MedGemma with $+8.16\%$ on VQA-RAD and $+9.12\%$ on MIMIC-Diff-VQA. For CheXagent on MIMIC-Diff-VQA, VGS-Decoding maintains baseline performance ($+0.12\%$), which is expected as CheXagent was specifically trained on MIMIC-CXR's official training set. Notably, VGS-Decoding is the only method that does not significantly degrade this well-calibrated model's performance, unlike VCD ($-4.37\%$), DoLA ($-2.77\%$), and OPERA ($-5.04\%$).
%%%%%%%%%%%%%%%%%%%%%%%%%%%%%%%%%%%%%
\begin{table*}[t]
\centering
\caption{Performance comparison on VQA-RAD (200 open + 251 closed) and 
MIMIC-Diff-VQA (8,380 open + 4,741 closed). Best results in \textbf{bold}. 
VGS-Decoding is the only method that consistently improves all models.}
\label{tab:main_results}
\resizebox{\textwidth}{!}{%
\begin{tabular}{ll|ccc|c|ccc|c}
\toprule
& & \multicolumn{4}{c|}{\textbf{VQA-RAD}} & 
  \multicolumn{4}{c}{\textbf{MIMIC-Diff-VQA}} \\
\textbf{Model} & \textbf{Method} & 
\textbf{Open} & \textbf{Closed} & \textbf{Overall} & $\boldsymbol{\Delta}$ & 
\textbf{Open} & \textbf{Closed} & \textbf{Overall} & $\boldsymbol{\Delta}$ \\
\midrule

\multirow{5}{*}{LLaVA-Med}
& Greedy       & 34.45 & 68.92 & 53.64 & --     & 28.04 & 48.39 & 35.39 & --     \\
& VCD          & 30.85 & 61.20 & 47.71 & -5.93  & 25.98 & 46.42 & 33.33 & -2.06  \\
& DoLA         & 32.76 & 58.96 & 47.34 & -6.30  & 28.75 & 47.94 & 35.68 & +0.29  \\
& OPERA        & 33.22 & 61.69 & 49.05 & -4.59  & 21.18 & 46.02 & 30.14 & -5.25  \\
\rowcolor{green!15}
& VGS (Ours)   & \textbf{38.90} & \textbf{72.91} & \textbf{57.75} & \textbf{+4.11} 
               & \textbf{30.84} & \textbf{55.79} & \textbf{39.86} & \textbf{+4.47} \\
\midrule

\multirow{5}{*}{CheXagent}
& Greedy       & 22.02 & 70.92 & 49.24 & --     & 44.06 & 82.07 & 57.79 & --     \\
& VCD          & 21.73 & 68.53 & 47.78 & -1.46  & 38.88 & 79.14 & 53.42 & -4.37  \\
& DoLA         & 20.73 & 68.92 & 47.55 & -1.69  & 39.78 & 81.94 & 55.02 & -2.77  \\
& OPERA        & 20.50 & 69.32 & 47.67 & -1.57  & 36.19 & 82.03 & 52.75 & -5.04  \\
\rowcolor{green!15}
& VGS (Ours)   & \textbf{23.48} & \textbf{71.31} & \textbf{50.10} & \textbf{+0.86} 
               & 43.99 & \textbf{82.53} & \textbf{57.91} & \textbf{+0.12}                            \\ %43.99%	82.53%	57.93%
\midrule

\multirow{5}{*}{MedGemma}
& Greedy       & 49.50 & 61.75 & 56.32 & --     & 25.97 & 73.55 & 43.16 & --     \\
& VCD          & 50.29 & 57.77 & 54.45 & -1.87  & 29.38 & 68.76 & 43.61 & +0.45  \\
& DoLA         & 51.91 & 72.51 & 63.38 & +7.06  & 32.56 & 76.82 & 48.55 & +5.39  \\
& OPERA        & 48.90 & 65.74 & 58.27 & +1.95  & 28.35 & 75.53 & 45.40 & +2.24  \\
\rowcolor{green!15}
& VGS (Ours)   & \textbf{53.05} & \textbf{73.58} & \textbf{64.48} & \textbf{+8.16} 
               & \textbf{34.95} & \textbf{82.91} & \textbf{52.28} & \textbf{+9.12} \\
\bottomrule

\end{tabular}%
}
\end{table*}
%%%%%%%%%%%%%%%%%%%%%%%%%%%%%%%%%%%%%%%%S
% \subsubsection{Consistency Analysis}

% Table~\ref{tab:consistency} summarizes the consistency of each method across all experiments. VGS-Decoding achieves improvements in 5 out of 6 model-dataset combinations (83\% consistency), substantially outperforming VCD (17\%), OPERA (33\%), and DoLA (50\%).

% \begin{table}[t]
% \centering
% \caption{Consistency analysis across model-dataset combinations.}
% \label{tab:consistency}
% \begin{tabular}{lcc}
% \toprule
% Method & Wins & Consistency \\
% \midrule
% VCD & 1/6 & 17\% \\
% OPERA & 2/6 & 33\% \\
% DoLA & 3/6 & 50\% \\
% VGS (Ours) & \textbf{5/6} & \textbf{83\%} \\
% \bottomrule
% \end{tabular}
% \end{table}

% \subsubsection{Statistical Significance}

% Table~\ref{tab:significance} reports statistical significance using McNemar's test with 95\% bootstrap confidence intervals on VQA-RAD. All improvements are statistically significant, with LLaVA-Med and MedGemma achieving $p<0.001$.
%%%%%%%%%%%%%%%%%%%%%%%%%%%%%%%%%%%%%%%%%%%%%%
% \begin{table}[t]
% \centering
% \caption{Statistical significance of VGS-Decoding on VQA-RAD.}
% \label{tab:significance}
% \begin{tabular}{lccc}
% \toprule
% Model & Overall & 95\% CI & $p$-value \\
% \midrule
% LLaVA-Med & 57.75\% & [53.2, 62.3] & $<$0.001 \\
% CheXagent & 50.10\% & [45.5, 54.8] & 0.038 \\
% MedGemma & 64.48\% & [59.9, 69.0] & $<$0.001 \\
% \bottomrule
% \end{tabular}
% \end{table}
%%%%%%%%%%%%%%%%%%%%%%%%%%%%%%%%%%%%%%%%%%%%%S

\begin{table}[t]
\centering
\caption{Component ablation study of VGS-Decoding on VQA-RAD (LLaVA-Med). 
Each row adds one component incrementally.}
\label{tab:ablation_component}
\begin{tabular}{lcccc}
\toprule
\textbf{Configuration} & \textbf{Open} & \textbf{Closed} & \textbf{Overall} & $\boldsymbol{\Delta}$ \\
\midrule
Baseline (Greedy)          & 34.45 & 68.92 & 53.64 & --     \\
+ Fixed weight (VCD-style) & 30.85 & 61.20 & 47.71 & \textcolor{red}{-5.93}  \\
+ VGS metric only          & 35.49 & 69.32 & 54.32 & +0.68  \\
\rowcolor{green!15}
+ VGS + Mult.\ reweight    & \textbf{38.90} & \textbf{72.91} & \textbf{57.75} & \textbf{+4.11} \\
\bottomrule
\end{tabular}
\end{table}
%%%%%%%%%%%%%%%%%%%%%%%%%%%%%%%%%%%%%%%%%
%%%%%%%%%%%%%%%%%%%%%%%%%%%%%%%%%%%%%%%%%%%%
\begin{table}[!htb]
\centering
\caption{Effect of reweighting strength $\alpha$ on VQA-RAD. 
$\alpha{=}0.0$ corresponds to greedy decoding. 
Optimal values per model are in \textbf{bold}.}
\label{tab:ablation_alpha}
\begin{tabular}{c|cc|cc}
\toprule
& \multicolumn{2}{c|}{\textbf{LLaVA-Med}} & 
  \multicolumn{2}{c}{\textbf{MedGemma}} \\
$\boldsymbol{\alpha}$ & \textbf{Overall} & $\boldsymbol{p}$\textbf{-value} 
                      & \textbf{Overall} & $\boldsymbol{p}$\textbf{-value} \\
\midrule
0.0 & 53.64          & --     & 56.32          & --       \\
0.5 & 56.95          & 0.046  & 59.00          & 0.002    \\
\rowcolor{green!15} 1.0 & 57.75          & 0.025  & 64.48          & 0.001    \\
1.5 & \textbf{58.79} & 0.001  & 64.37          & $<$0.001 \\
2.0 & 57.50          & 0.014  & \textbf{65.90} & $<$0.001 \\
\bottomrule
\end{tabular}
\end{table}

\subsubsection{Ablation Study}
% \begin{table}[ht]
% \centering
% \caption{Ablation study of VGS-Decoding components on VQA-RAD (MedGemma)}
% \label{tab:ablation_medgemma}
% \begin{tabular}{lcccc}
% \hline
% \textbf{Configuration} & \textbf{Open} & \textbf{Closed} & \textbf{Overall} & \textbf{$\Delta$} \\
% \hline
% Baseline  & 49.50\% & 61.75\% & 56.32\% & - \\
% + Fixed weight (VCD-style) & 50.29\% & 57.77\% & 54.45\% & -1.87\% \\
% + VGS metric only & 49.59\% & 62.20\% & 56.60\% & 0.28\% \\
% + VGS + Multi. reweight & 53.05\% & 73.58\% & 64.48\% & 8.16\% \\ %Multiplicative 
% \hline
% \end{tabular}
% \end{table}

%%%%%%%%%%%%%%%%%%%%%%%%%%%%%%%%%%%%%

%%%%%%%%%%%%%%%%%%%%%%%%%%%%%%%%%%%%%%%%
Table~\ref{tab:ablation_component} analyzes the contribution of each component in VGS-Decoding. VCD-style fixed-weight contrastive decoding degrades performance by $-5.93\%$, confirming that existing methods fail on medical VQA. Computing VGS alone without reweighting provides modest improvement of $+0.68\%$, demonstrating that the metric captures useful signal. The full VGS-Decoding with multiplicative reweighting achieves the best results with $+4.11\%$ improvement, showing that both the VGS metric and adaptive reweighting are necessary.

Table~\ref{tab:ablation_alpha} examines the effect of reweighting strength $\alpha$. All non-zero values achieve statistically significant improvements over baseline ($p<0.05$). Optimal $\alpha$ varies by model, with 1.5 for LLaVA-Med and 2.0 for MedGemma, but $\alpha=1.0$ provides consistent improvements across all models and is used as default.

\section{Conclusion}
\label{sec:conclusion}
We proposed VGS-Decoding, a training-free method that mitigates 
hallucinations in medical VLMs by adaptively reweighting token 
probabilities via the Visual Grounding Score (VGS), which exploits 
the observation that hallucinated tokens are resilient to visual 
degradation while grounded tokens are not. Experiments on VQA-RAD 
and MIMIC-Diff-VQA across LLaVA-Med, CheXagent, and MedGemma 
demonstrate that VGS-Decoding is the only method that consistently 
improves all models, achieving up to $+9.12\%$ overall improvement 
with only $2\times$ inference overhead and no additional training.

%
% ---- Bibliography ----
%
% BibTeX users should specify bibliography style 'splncs04'.
% References will then be sorted and formatted in the correct style.
%
\bibliographystyle{splncs04}
\bibliography{references}
%
% \begin{thebibliography}{8}
% \bibitem{ref_article1}
% Author, F.: Article title. Journal \textbf{2}(5), 99--110 (2016)

% \bibitem{ref_lncs1}
% Author, F., Author, S.: Title of a proceedings paper. In: Editor,
% F., Editor, S. (eds.) CONFERENCE 2016, LNCS, vol. 9999, pp. 1--13.
% Springer, Heidelberg (2016). \doi{10.10007/1234567890}

% \bibitem{ref_book1}
% Author, F., Author, S., Author, T.: Book title. 2nd edn. Publisher,
% Location (1999)

% \bibitem{ref_proc1}
% Author, A.-B.: Contribution title. In: 9th International Proceedings
% on Proceedings, pp. 1--2. Publisher, Location (2010)

% \bibitem{ref_url1}
% LNCS Homepage, \url{http://www.springer.com/lncs}, last accessed 2023/10/25
% \end{thebibliography}
\end{document}